\documentclass[runningheads]{llncs}
\usepackage[T1]{fontenc}
\usepackage{graphicx}
\usepackage{amsmath}
\usepackage{marginnote}
\usepackage{bbding}
\usepackage{cite}
\usepackage[rgb,x11names]{xcolor}
\usepackage{amssymb}
\usepackage{amsmath}
\usepackage{subfig}
\usepackage{hyperref}
\hypersetup{colorlinks=true,citecolor=blue}
\usepackage{graphicx}
\usepackage[]{algorithm2e}
\usepackage{mathrsfs}
\usepackage{bbding}
\usepackage{rotating}
\usepackage{multirow}
\usepackage{marginnote}
\usepackage{overpic}
\usepackage{colortbl}
\usepackage[labelfont=bf]{caption}
\usepackage{bbding}
\usepackage{tabularx}
\usepackage[marginal]{footmisc}

\begin{document}
\title{Mask-TS Net: Mask Temperature Scaling Uncertainty Calibration for Polyp Segmentation}
%
%

\author{Yudian Zhang\inst{*}\and
Chenhao Xu\inst{*}\and
Kaiye Xu\inst{}\and
Haijiang Zhu\inst{(}\Envelope\inst{)}}
\authorrunning{Zhang et al.}
\titlerunning{Mask-TS Uncertainty Calibration for Segmentation}%
%
\institute{CIST, Beijing University of Chemical Technology, Beijing 100029, China \\
\email{\{zyd, xuch, xuky\}@buct.edu.cn}\\
\email{zhuhj@mail.buct.edu.cn}
}
\maketitle              
\footnote{* means the authors contributed equally to this work and should be considered co-first authors.\\}
\begin{abstract}

Lots of popular calibration methods in medical images focus on classification, but there are few comparable studies on semantic segmentation. In polyp segmentation of medical images, we find most diseased area occupies only a small portion of the entire image, resulting in previous models being not well-calibrated for lesion regions but well-calibrated for background, despite their seemingly better Expected Calibration Error (ECE) scores overall. Therefore, we proposed four-branches calibration network with Mask-Loss and Mask-TS strategies to more focus on the scaling of logits within potential lesion regions, which serves to mitigate the influence of background interference. In the experiments, we compare the existing calibration methods with the proposed Mask Temperature Scaling (Mask-TS). The results indicate that the proposed calibration network outperforms other methods both qualitatively and quantitatively.
\end{abstract}
\section{Introduction}
With the rapid development of deep learning technology, neural networks are increasingly widely used in the field of medical image processing. Despite deep learning has promise in medical imaging, concerns over reliability and trust impede their implementation in real-world clinical contexts, where any misjudgment will bring great risks to the doctor's diagnostic process and the patient's treatment. Ideal semantic segmentation networks should not merely excel in precision, but they should also possess the capability to indicate when and where their predictions may be unreliable or prone to error. If the segmentation network shows high uncertainty of its predictions, a medical expert is needed to double-check such doubtful regions. In order to achieve this goal, uncertainty estimation is essential.

However, the probabilities output from the semantic segmentation model are often overconfident because of overfitting~\cite{overconf1, overconf2, ts}, which may mislead the final decision of doctors. To solve this, lots of calibration techniques~\cite{asurvey, areview} are proposed including regularization methods like data augmentation~\cite{augmentation1, augmentation2} and uncertainty estimation approaches like ensemble of neural networks~\cite{Ensemble1, Ensemble2}. However, these approaches can potentially compromise the segmentation model's accuracy. So, our focus lies in post-hoc calibration methodologies, which build an independent post-processing model separated from the prediction model. This approach can improve probabilistic calibration while keeping the model's original prediction unchanged.

What is more, the majority of existing post-hoc probability calibration methods are originally designed for image classification and typically yield a solitary class probability per entire image~\cite{ps,ts,ets} while as for segmentation pixel-wised probabilities of each pixel are called for. So these methods cannot be directly applied to medical image segmentation. Although there are some methods for semantic segmentation, the difference between background and target in the calibration process is ignored.

Our goal is to develop a post-hoc calibration method for two-label medical image semantic segmentation, which pay attention to the distinction between background and target in calibration process and retains label prediction and a model’s segmentation accuracy. Then, an intuitive uncertainty map with highly practical application value for clinical contexts is given.

Here are our specific contributions: 
\begin{itemize}
    \item Prediction-based mask for temperature scaling is proposed to finely select the region of interest, which truly calibrates the probabilities of lesion area.
    \item For segmentation task, pixel-wised probabilities of each pixel are produced and spatial relations of pixels are concerned via convolutional neural networks.
    \item The probability calibration of two-label semantic segmentation problem is realized without compromising the original prediction accuracy.
\end{itemize}
\section{Related Work}
The most significant feature of the post-calibration method is that it makes the out-put probability closer to the true probability distribution without changing the prediction results of the original network, which makes the effect of the model better. Using post-calibration methods to process the output of an already trained network, such as Platting Scaling\cite{ps}, which combines vector machines with Sigmoid, yields better regularized maximum likelihood estimates. In the field of classification, one of the most basic but effective methods is Temperature Scaling (TS)\cite{ts}. It divides the logits by a constant $T$ so that the output of the model is calibrated. By extension, there is Ensemble Temperature Scaling (ETS)\cite{ets}. Although TS performs well in classification tasks, for semantic segmentation tasks, the positional relationship between pixels needs to be taken into account and probability calibration for the each pixels is needed. As an extension of TS, Local Temperature Scaling (LTS)\cite{lts} considers that the parameter T of each pixel varies due to the relative positions. But it ignores the difference between background and target. As a result, it is not well-calibrated for lesion regions but well-calibrated for background, despite their seemingly high ECE scores. Factually, these methods can benefit from our Mask-Loss and Mask-TS strategies to more target at regions of interest and calibrate corresponding probabilities.

\begin{figure}[t]
\includegraphics[width=\textwidth]{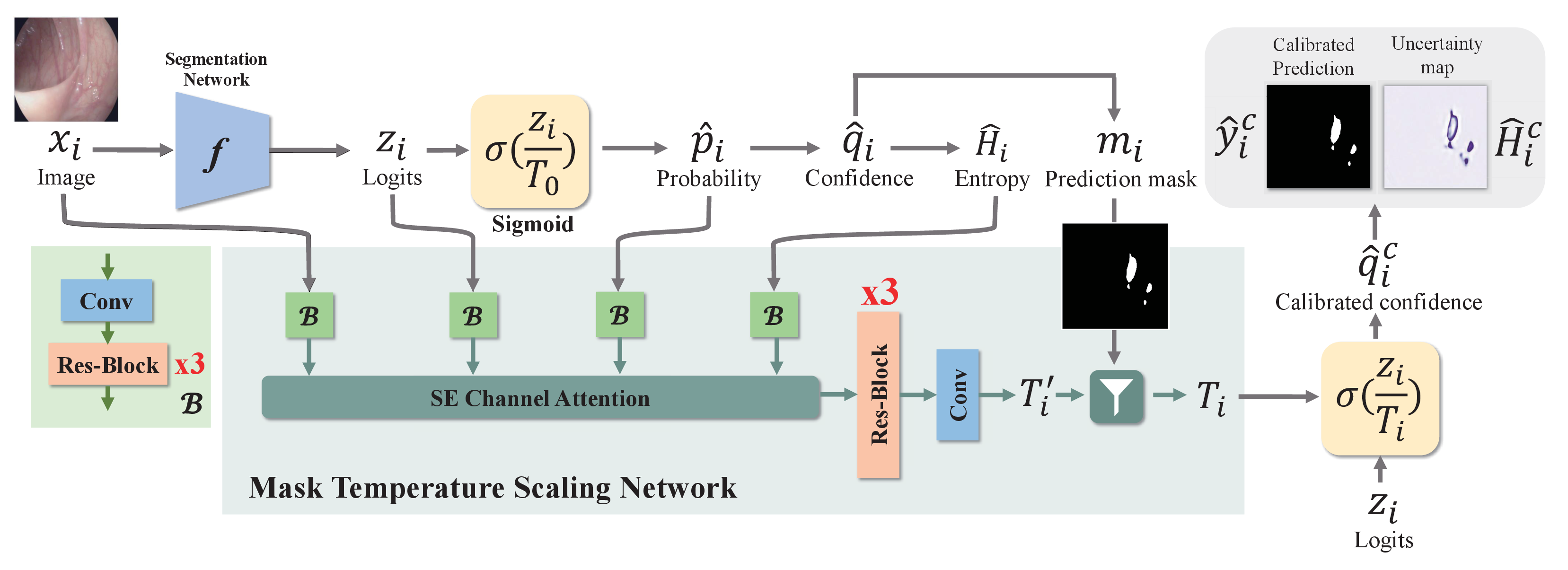}
\caption{Structure of Mask-TS Network.} \label{calinet}
\end{figure}

\section{Calibration Network}

The existing probabilistic calibration methods, such as platting scaling, TS, and ETS primarily focus on classification tasks rather than semantic segmentation tasks. Considering the spatial relations of pixels in an image, image segmentation should not be simply treated as classification. We therefore design a Mask-TS calibration network that performs different pre-temperature scaling on each pixel via convolutional networks. And then, a mask based on prediction is applied to the pre-temperature scaled parameters in order to pay more attention to the potential diseased area. Here are the details:

\subsection{Four Branches Calibration Network}\label{fourbranch}
The predictive confidence of the two labels $\hat{q}_i$ is calculated by Eq.~(\ref{eq1}) because our task is two-label segmentation and  the original probability $\hat{p}_i$ is the original sigmoid output. Original prediction mask $m_i$, namely $\hat{y}_i$, is similarly obtained like Eq. (\ref{eq2}) where $\hat{q}^c_i$ is the calibrated confidence and $\hat{y}^c_i$ is the calibrated prediction.

\begin{equation}
\hat{q}_i=\left \{ \begin{matrix}
\hat{p}_i&, \text{if}\ \hat{p}_i\geq0.5 \\
1-\hat{p}_i&, \text{otherwise}\\
\end{matrix}\right.
\label{eq1}
\end{equation}

\begin{equation}
\hat{y}^c_i=\left
\{\begin{matrix}
 1 &, \text{if}\  \hat{q}^c_i\ge0.5 \\
 0 &, \text{otherwise}
\end{matrix}\right.
\label{eq2}
\end{equation}

As shown in Fig.~\ref{calinet}, the original image $x_i$, model output logits $z_i$, probability map $\hat{p}_i$ and uncertainty map $\hat{H}_i$ are separately input into network block $\mathcal{B}$ composed of convolution and residual neural networks, which means taking into consideration the spatial relationship between pixels. Then information from the four branches are adaptively weighted and combined under the action of a channel attention layer.

Our work is inspired by temperature scaling (TS)\cite{ts} for classification probability calibration, and improved TS\cite{its} . However, the key component referred to as the shape prior network in improved TS, which requires higher training costs for a denoising autoencoder, is not utilized in our work. To avoid this, we creatively input $\hat{H}_i$ and $\hat{p}_i$ into proposed calibration network. We empirically found that our method yields comparable results to the improved TS. These two branches respectively play a role in enhancing the edge shape and treating the predicted truth value more carefully while $x_i$ contains rich original information and $z_i$ possesses segmentation-related features because it originates from a segmentation network. 

We train the post-hoc calibration network using the binary cross entropy loss Eq.~(\ref{eq9}) for two-label image segmentation, even though negative log likelihood loss\cite{nll} is commonly used for multi-label task. Uncertainty map is used to quantify the uncertainty of pre-dictive segmentation using the aleatoric uncertainty \cite{aleatoricuncertainty,entropy} for each pixel. For two-label segmentation task, this is measured by the entropy of the confidence Eq.~(\ref{eq10}).

\begin{small}
\begin{equation}
    \mathcal{L}_c=\frac{-1}{H\cdot W }\sum_{u,v=1,1 }^{H,W}\left \{ y_i(u,v)\cdot\text{log} [\sigma(\frac{z_i(u,v)}{T_i(u,v)})]+[1-y_i(u,v) ]\cdot\text{log}[1-\sigma(\frac{z_i(u,v)}{T_i(u,v)})] \right \}
    \label{eq9}
\end{equation} 
\end{small}

\begin{equation}
\hat{H}_i=\mathcal{H}(\hat{q}_i)=-[\hat{q}_i\text{log}_2(\hat{q}_i)+(1-\hat{q}_i)\text{log}_2(1-\hat{q}_i)]
\label{eq10}
\end{equation}

\begin{figure}[t]
\includegraphics[width=\textwidth]{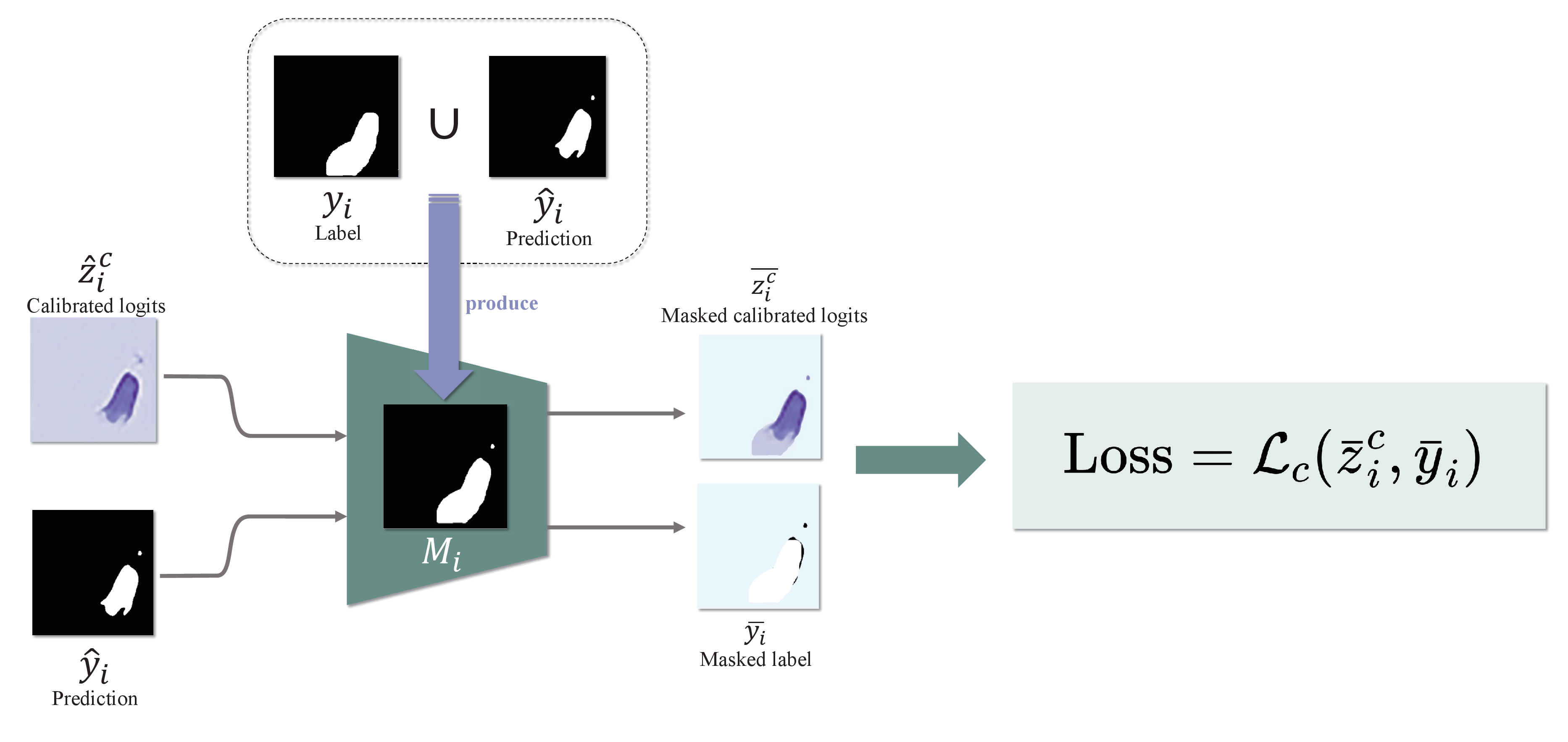}
\caption{Workflow of the proposed Mask-Loss method.} \label{maskloss}
\end{figure}

\subsection{Mask-Loss}
In practice, doctors more focus on positive predictions that may indicate potentially diseased area. Nevertheless, we find that most diseased area occupies only a small portion of the entire image, resulting in less important background predictions dominating the calculation of ECE. It also leads to previous models not being well-calibrated for the actual lesion regions, despite their high ECE scores. This is detailed in section~\ref{Evaluation }. 

So Mask-Loss strategy is proposed during the training stage. Specifically, as shown in Eq.~(\ref{eq3})(\ref{eq4})(\ref{eq5}), when computing BCE-Loss, we first perform a logical OR operation on the positive labels and the positive predictions to obtain the mask $M_i$. Then masked and calibrated logits $\bar{z}_i^c$ and masked label $\bar{y}_i$ , both processed through $M_i$, are used to calculate BCE-Loss $\mathcal{L}_c$ using Eq.~(\ref{eq9}). Fig.~\ref{maskloss} also shows the workflow of Mask-Loss and it should be noted that the light blue areas, seen as background, in $\bar{z}_i^c$ and $\bar{y}_i$ are excluded from the loss calculation. Thus, the background interference is weakened while positive predictions are more focused and concerned.

\begin{equation}
M_i =({y_i}==1) \text{or} (\hat{y_i} ==1)
\label{eq3}
\end{equation}

\begin{equation}
    \bar{z^c_i}=z^c_i\odot M_i=(\frac{z_i}{T_i})\odot M_i
    \label{eq4}
\end{equation}

\begin{equation}
\bar{y_i} =y_i\odot M_i
\label{eq5}
\end{equation}

\begin{figure}[t]
\includegraphics[width=\textwidth]{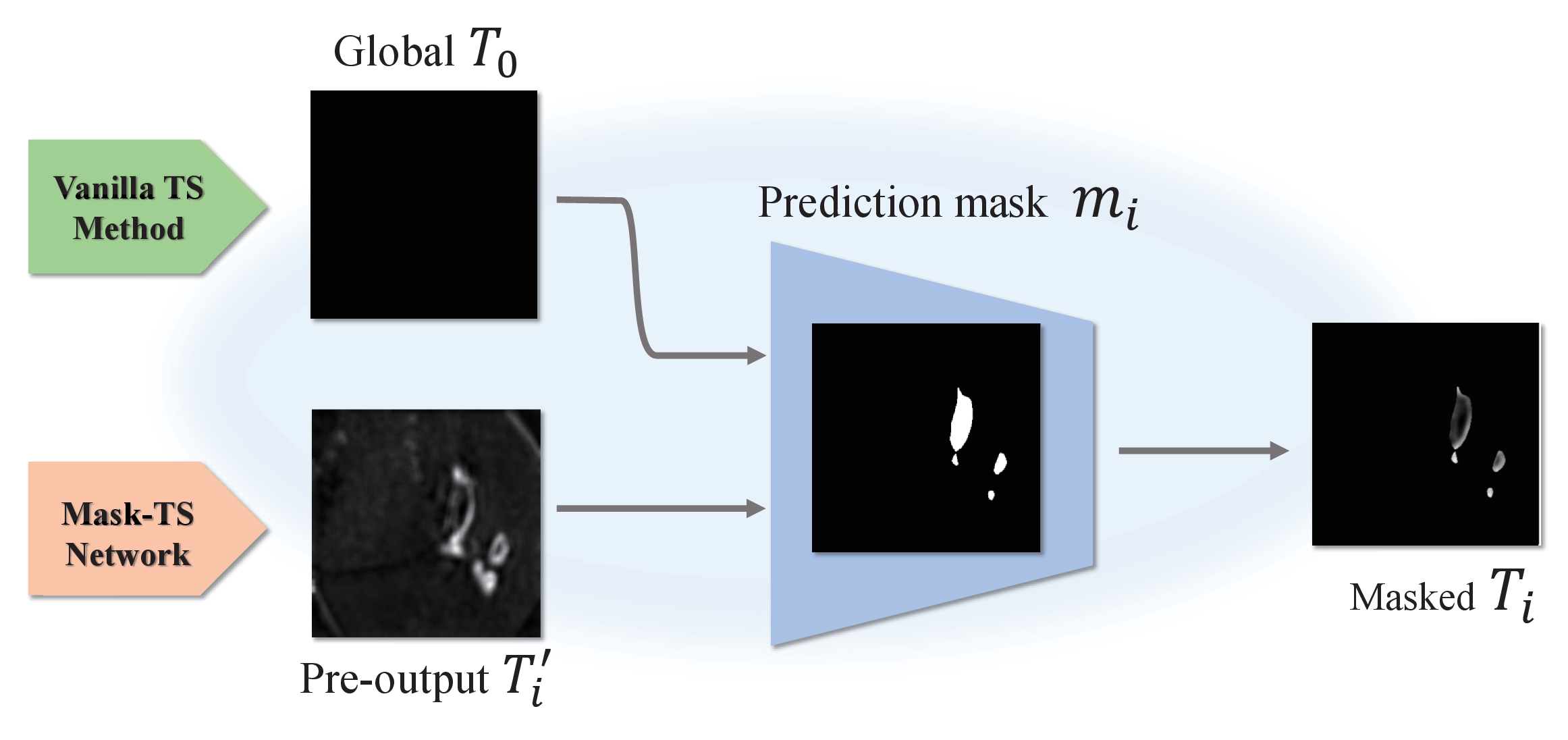}
\caption{Architecture for Mask-TS.} \label{maskts}
\end{figure}

\subsection{Mask-TS}
To avoid the lack of constraints on the background pixels from causing an increase of the temperature parameters and pay attention to the calibration of potential lesion areas, we proposed the prediction-based Mask-TS method.

Through Mask-TS, for predictions classified as background, the corresponding temperature parameter $T_i$ is set as the global temperature parameter $T_0$ produced by vanilla TS, while for predictions classified as lesion areas, $T_i$ remains as the pre-output temperature parameter $T'_i$ of the calibration network. Eq.~(\ref{eq6}) shows how to obtain a masked temperature map $T_i$ with the mathematical expression. Fig.~\ref{maskts} shows the architecture for Mask-TS. It illustrates that $T'_i$ may miscalibrate the background areas due to the lack of constraints caused by Mask-Loss and the masked $T_i$ more focus on potential lesion areas while eliminate the background miscalibration.

Through practice, it is found that the final prediction of the model remains unchanged even $z_i$ is scaled in different forms. The reason is that we use Eq.~(\ref{eq1})(\ref{eq2}) resulting in the area where $\hat{p}_i$ is greater than 0.5 still maintain its original confidence value above 0.5 even after different scaling operations on $z_i$ and keep the prediction results unchanged, creating the conditions for confidence calibrating without changing the original prediction results.

\begin{equation}
T_i=\left\{\begin{matrix}
 T_i'&, \text{if}\ \hat{y}_i=1\\
 T_0&, \text{otherwise}
\end{matrix}\right.
\label{eq6}
\end{equation}

\begin{table}[!ht]
   \caption{Dataset allocation. }\label{tab1}
    \centering
    \begin{tabular}{|c|c|c|c|}
    \hline
        Dataset& Image Num & Segmentation Net & Calibration Net   \\ \hline
        CVC-ClinicDB\cite{cvccli} & 550 & Train & - \\ 
        Kvasir\cite{kvasir} & 900 & Train & - \\ 
        CVC-ColonDB\cite{cvccol} & 380 & - & Train\\ 
        CVC-300\cite{cvc300} & 60 & - & Validate \\ 
        ETIS-LaribPolypDB\cite{etis} & 196 & - & Test \\ \hline
    \end{tabular}
\end{table}

\section{Experiment}
We show the calibration performance of proposed method on ETIS-LaribPolypDB\cite{etis} dataset and also show the practical uncertainty maps. We compare our Mask-TS with baseline methods such as TS, ETS and LTS and conduct ablation studies based on four branches of our calibration network. As shown in Table \ref{tab1}, for calibration part, CVC-ColonDB\cite{cvccol} is used as training dataset, CVC-300\cite{cvc300} as validation dataset and ETIS-LaribPolypDB as test dataset. The size of all the images is adjusted to 352x352 and Adam optimizer is used with a learning rate of 1e-4 and 200 epochs.

\subsection{Evaluation Metrics}
To measure the performance of probability calibration, we use the following metrics. Qualitatively, uncertainty map compared with error map is used to visually evaluate calibration results. Quantitatively, reliability diagram \cite{rd,rd2}, expected calibration error \cite{ece} (ECE), maximum calibration error \cite{ece} (MCE), static calibration error \cite{SCEACE} (SCE), and adaptive calibration error \cite{SCEACE} (ACE) are used. These metrics, originally devised for classification, are adjusted for semantic segmentation by treating each pixel's confidence as a separate sample. Computations are performed across 10 evenly-spaced bins.

When calculating quantitative metrics, method local evaluation regions is used for testing. This plays a role in excluding the influence of less important background in practical application and more focusing on calibration performance of potential lesion area. In practice, for each prediction map to be tested, the center point is randomly selected in the height and width range of (70, 352-70), and then 10 small patches (72x72) are taken based on the location of the center point.

\begin{figure}[t]
\includegraphics[width=\textwidth]{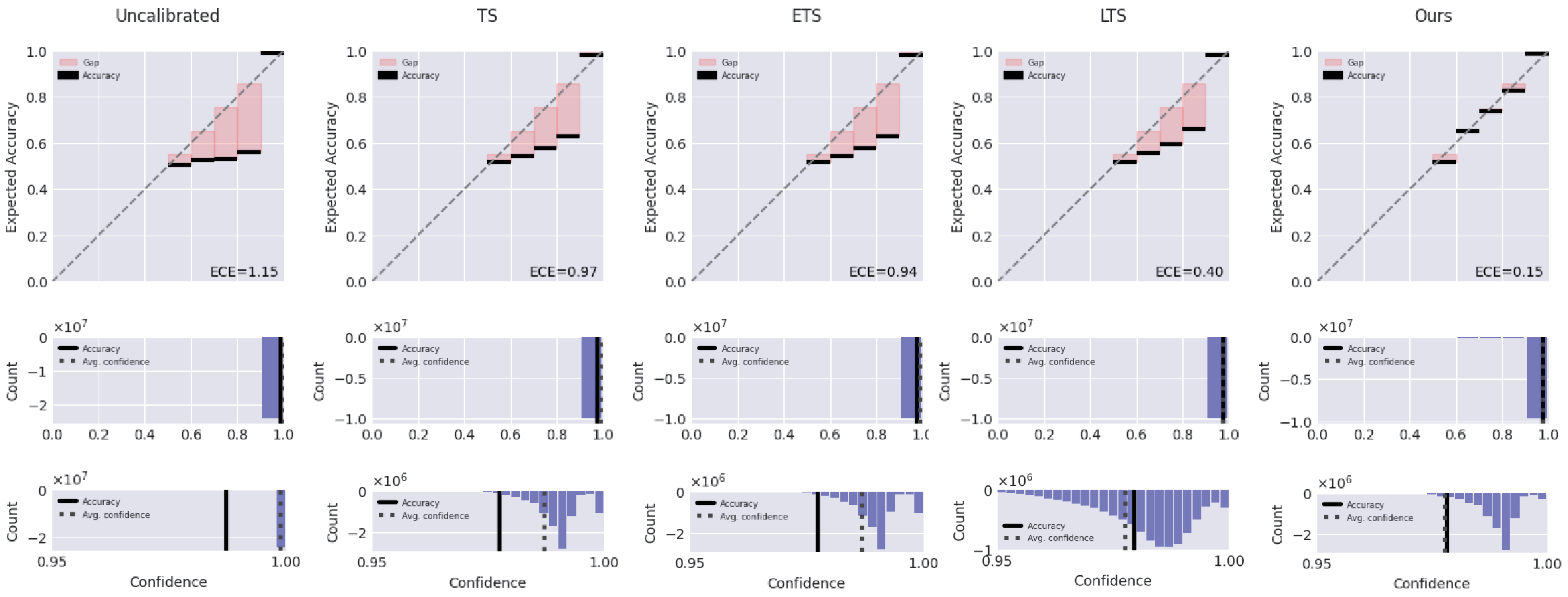}
\caption{Reliability diagrams and confidence histograms of comparative experiment.} \label{chart}
\end{figure}

\begin{figure}[t]
\includegraphics[width=\textwidth]{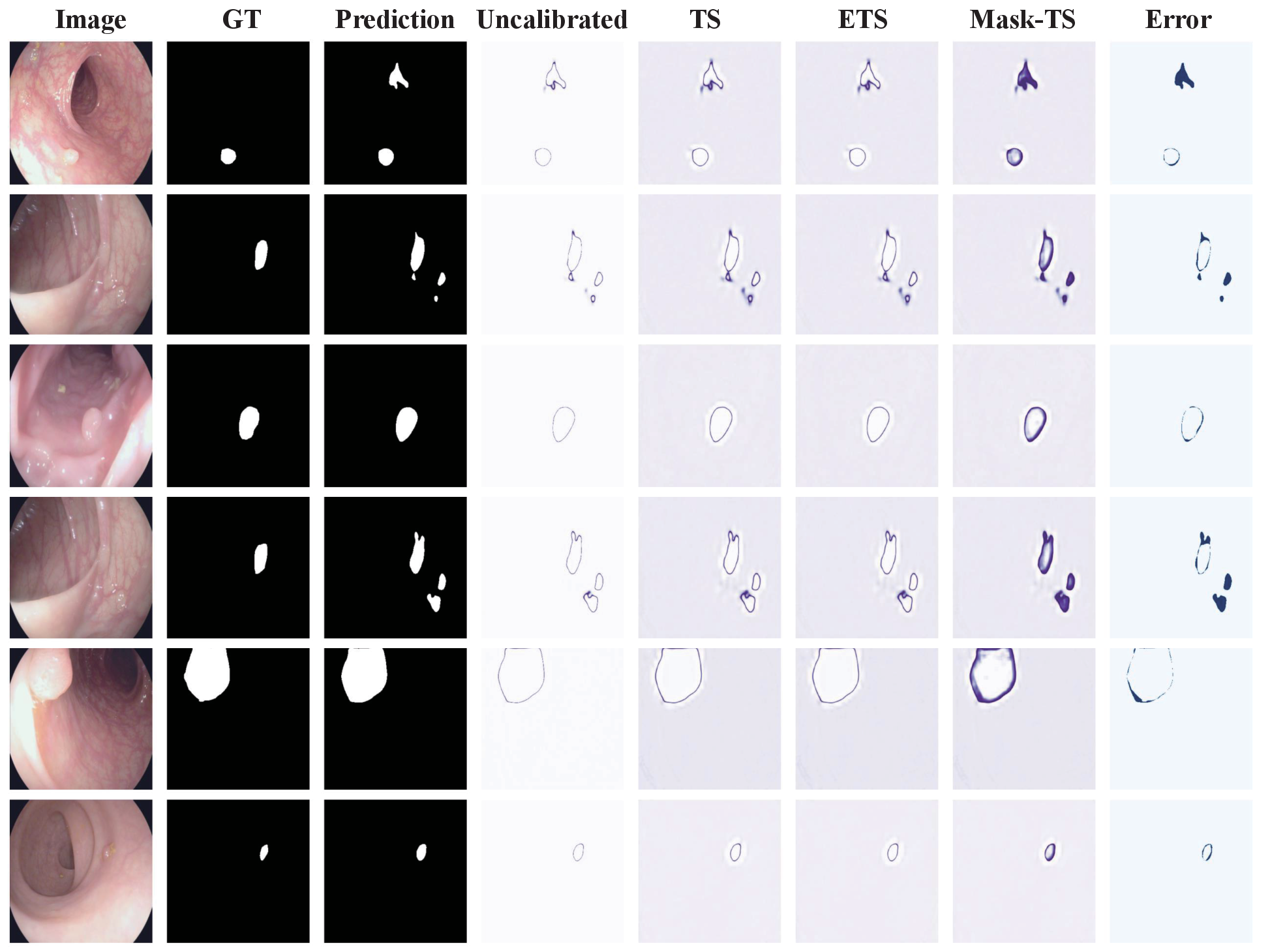}
\caption{Uncertainty maps of comparative experiment.} \label{result}
\end{figure}

\subsection{Evaluation and Results}\label{Evaluation }
As shown in Fig.~\ref{chart} , the top row shows our calibration method achieves the minimal misalignment (red gap) between confidence and accuracy. The middle and bottom rows of the confidence histogram illustrate that the proposed method corrects over-confident predictions compared with other results.

Note that there is a common problem of overconfident miscalibration in image segmentation \cite{overconf1, overconf2}.  In our task, each pixel is given a probability by Eq.~(\ref{eq1}). Because of this either-or mechanism where the confidences of positive and negative samples sum to one, overconfident miscalibration is greatly amplified. This leads to a very high confidence (close to 0.99 in our task) for non-boundary pixels with a large proportion, which has a large impact on calibration effect. This can also explain why there are many samples between the confidence interval [0.9,1) in the histograms.

Therefore, we find that in the task of two-label image segmentation calibration, although the previous models such as TS, ETS, and LTS reduce the value of ECE, they actually only reduce the ECE of the background area and do not really correct the confidence of the lesion area, nor can they give a calibration uncertainty map with reference value. This is why Mask-Loss and Mask-TS are needed so as to pay more attention to the lesion area. The Fig.~\ref{chart} and Table \ref{tab0} shows our model really makes the lesion area and the whole image well calibrated.
\begin{table}[!ht]
    \caption{Quantitative metrics results for 4 different calibration methods.}\label{tab0}
    \centering
    \begin{tabular}{|c|*{5}{c|}}
    \hline
        \makebox[0.1\textwidth][c]{Methods}  
        & \makebox[0.16\textwidth][c]{Uncalibrated} 
        & \makebox[0.16\textwidth][c]{TS} 
        & \makebox[0.16\textwidth][c]{ETS} 
        & \makebox[0.16\textwidth][c]{LTS} 
        &\makebox[0.16\textwidth][c]{OURS}  \\ \hline
        ECE \textdownarrow & 1.15 & 0.97 & 0.94 & 0.40 & \textbf{0.15}   \\ 
        MCE \textdownarrow & 29.60 & 22.82 & 22.74 & 19.62 & \textbf{3.30}  \\ 
        SCE \textdownarrow & 8.13 & 7.24 & 7.25 & 6.36 & \textbf{2.17} \\ 
        ACE \textdownarrow & 8.12 & 7.31 & 7.31 & 6.40 & \textbf{2.35}  \\ \hline
    \end{tabular}
\end{table}

As shown in Fig.~\ref{result}, the results with some mispredictions are specifically selected to test whether the uncertainty map is well calibrated. Comparison of the error and the uncertainty map shows that the doubt of the calibration model agrees well with the factual segmentation error.

\begin{figure}[!ht]
\includegraphics[width=\textwidth]{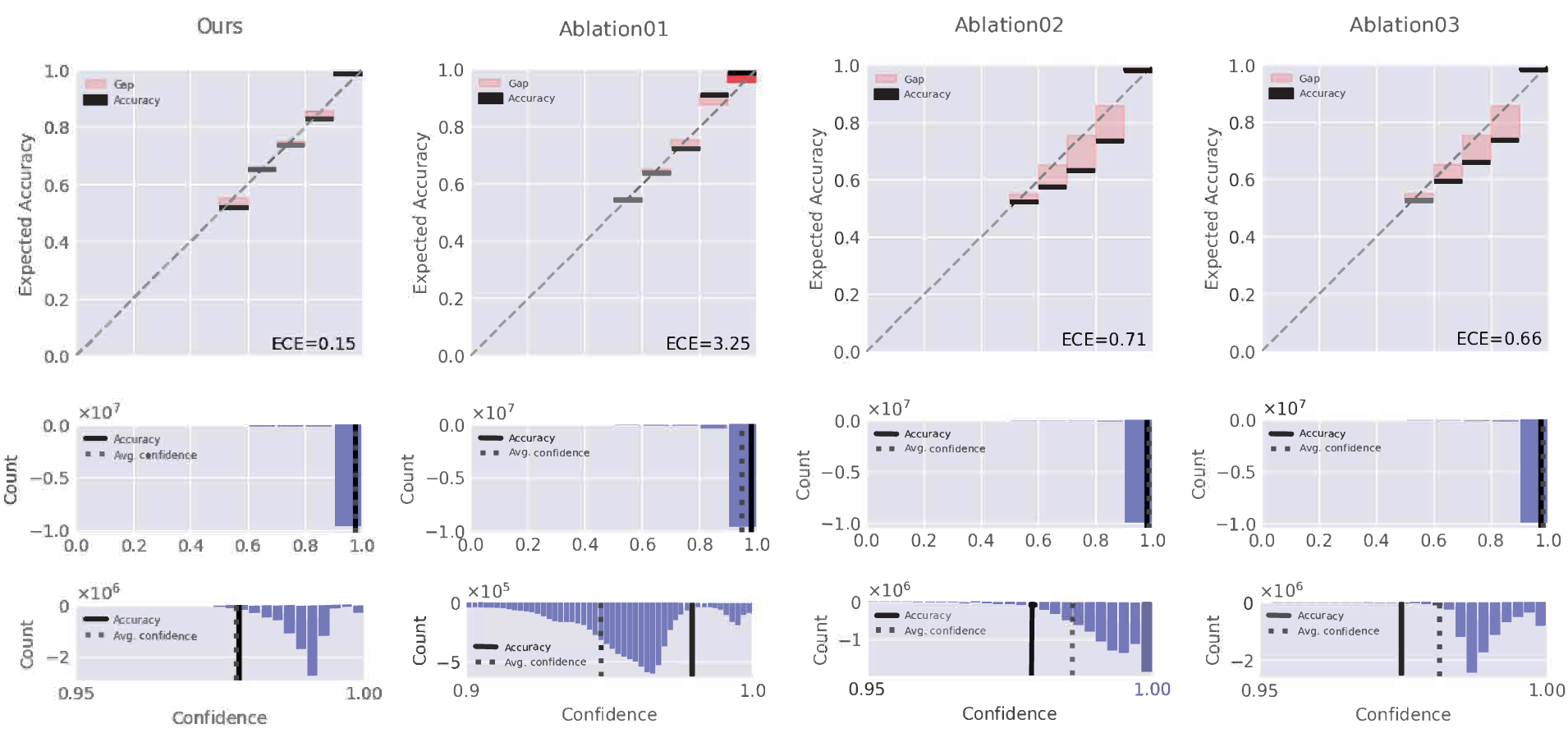}
\caption{Ablation results of reliability diagram \& confidence histogram.} \label{abchart}
\end{figure}

\begin{figure}[!ht]
\includegraphics[width=\textwidth]{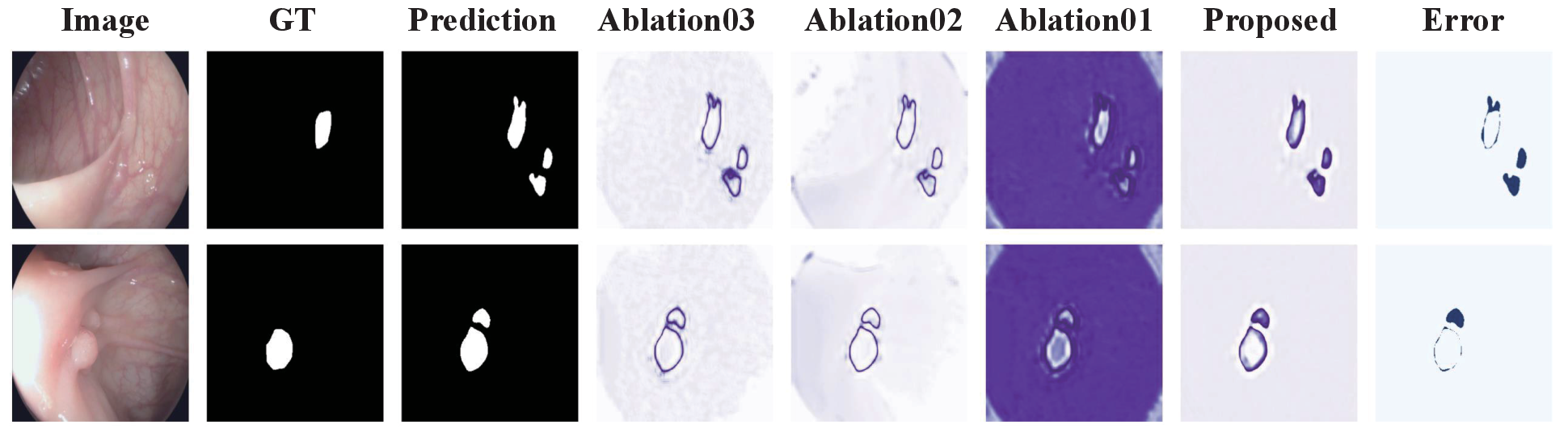}
\caption{Ablation results of uncertainty map.} \label{abresult}
\end{figure}

\subsection{Ablation studies}
As mentioned in section \ref{fourbranch}, our Calibration Network primarily consists of four branches: uncertainty maps from the vanilla TS, probabilities from vanilla TS, output logits and input image, which are concatenated through an attention mechanism followed by process of Mask-Loss and Mask-TS. In the ablation experiments, we ablate some key components: Ablation01 removes Mask-TS, Ablation02 eliminates Mask-Loss \& Mask-TS, and Ablation03 abolishes Mask-Loss \& Mask-TS \& probability from the vanilla TS. Subsequently, the calibration results are tested based on quantitative and qualitative metrics, as depicted in Fig.~\ref{abchart}, Fig.~\ref{abresult} and Table \ref{tab3}.
\begin{table}[!ht]
    \caption{Quantitative metrics results for ablation experiment.}\label{tab3}
    \centering
    \begin{tabular}{|c|*{4}{c|}}
    \hline
        \makebox[0.1\textwidth][c]{Methods}  
        & \makebox[0.18\textwidth][c]{OURS} 
        & \makebox[0.18\textwidth][c]{ABLATION01} 
        & \makebox[0.18\textwidth][c]{ABLATION02} 
        & \makebox[0.18\textwidth][c]{ABLATION03}\\ \hline
        ECE \textdownarrow& \textbf{0.15} & 3.25 & 0.71 & 0.66   \\ 
        MCE \textdownarrow& \textbf{3.30} & 3.84 & 12.33 & 11.98  \\ 
        SCE \textdownarrow& \textbf{2.17}& 7.06 & 5.45 & 5.70 \\ 
        ACE \textdownarrow& \textbf{2.35} & 7.06 & 5.44 & 5.73 \\ \hline
    \end{tabular}
\end{table}

In detail, the comparison of Ablation01 with Ablation02 and Ablation03 displays the calibration effect of Mask-Loss on the lesion area, where typically the region consists of a smaller proportion of pixels and exhibits lower confidence levels (0.5, 0.9), manifested as a reduction in the red gap. But it causes misalignment in background where confidence levels are higher, between (0.9, 1). Our method compared to Ablation01 illustrates the well calibrated effect among background area via introducing Mask-TS on overall network calibration. In a word, the calibration network achieves optimal performance when the four branches work together and the Mask-Loss \& Mask-TS strategies are adopted.

\section{Conclusion}
In this work, we propose a better calibration method for binary-label polyp image segmentation. Further work can be done to improve the accuracy by improving the segmentation network structure preferably with smaller original miscalibration, improving the calibration network structure to reduce miscalibration or using more detailed masks to reduce misalignment.

\subsubsection{Acknowledgements.} Research supported by the Training Program of Innovation and Entrepreneurship for Undergraduates in Beijing University of Chemical Technology (X202410010085).

%
%
%
\bibliographystyle{splncs04}
\bibliography{ref}

\begin{thebibliography}{10}
\providecommand{\url}[1]{\texttt{#1}}
\providecommand{\urlprefix}{URL }
\providecommand{\doi}[1]{https://doi.org/#1}

\bibitem{overconf2}
Bengio, Y., Goodfellow, I., Courville, A.: Deep learning, vol.~1. MIT press Cambridge, MA, USA (2017)

\bibitem{cvccli}
Bernal, J., S{\'a}nchez, F.J., Fern{\'a}ndez-Esparrach, G., Gil, D., Rodr{\'\i}guez, C., Vilari{\~n}o, F.: Wm-dova maps for accurate polyp highlighting in colonoscopy: Validation vs. saliency maps from physicians. Computerized medical imaging and graphics  \textbf{43},  99--111 (2015)

\bibitem{rd}
DeGroot, M.H., Fienberg, S.E.: Assessing probability assessors: calibration and refinement. Statistical decision theory and related topics III  \textbf{1},  291--314 (1982)

\bibitem{lts}
Ding, Z., Han, X., Liu, P., Niethammer, M.: Local temperature scaling for probability calibration. In: Proceedings of the IEEE/CVF International Conference on Computer Vision. pp. 6889--6899 (2021)

\bibitem{nll}
Friedman, J., Hastie, T., Tibshirani, R.: The elements of statistical learning. vol. 1 springer series in statistics. New York  (2001)

\bibitem{asurvey}
Gawlikowski, J., Tassi, C.R.N., Ali, M., Lee, J., Humt, M., Feng, J., Kruspe, A., Triebel, R., Jung, P., Roscher, R., et~al.: A survey of uncertainty in deep neural networks. Artificial Intelligence Review  \textbf{56}(Suppl 1),  1513--1589 (2023)

\bibitem{ts}
Guo, C., Pleiss, G., Sun, Y., Weinberger, K.Q.: On calibration of modern neural networks. In: International conference on machine learning. pp. 1321--1330. PMLR (2017)

\bibitem{overconf1}
Hastie, T., Friedman, J., Tibshirani, R., Hastie, T., Friedman, J., Tibshirani, R.: Kernel methods. The Elements of Statistical Learning: Data Mining, Inference, and Prediction pp. 165--192 (2001)

\bibitem{aleatoricuncertainty}
Kendall, A., Gal, Y.: What uncertainties do we need in bayesian deep learning for computer vision? Advances in neural information processing systems  \textbf{30} (2017)

\bibitem{Ensemble2}
Lakshminarayanan, B., Pritzel, A., Blundell, C.: Simple and scalable predictive uncertainty estimation using deep ensembles. Advances in neural information processing systems  \textbf{30} (2017)

\bibitem{entropy}
McClure, P., Rho, N., Lee, J.A., Kaczmarzyk, J.R., Zheng, C.Y., Ghosh, S.S., Nielson, D.M., Thomas, A.G., Bandettini, P., Pereira, F.: Knowing what you know in brain segmentation using bayesian deep neural networks. Frontiers in neuroinformatics  \textbf{13}, ~67 (2019)

\bibitem{Ensemble1}
Mehrtash, A., Wells, W.M., Tempany, C.M., Abolmaesumi, P., Kapur, T.: Confidence calibration and predictive uncertainty estimation for deep medical image segmentation. IEEE transactions on medical imaging  \textbf{39}(12),  3868--3878 (2020)

\bibitem{ece}
Naeini, M.P., Cooper, G., Hauskrecht, M.: Obtaining well calibrated probabilities using bayesian binning. In: Proceedings of the AAAI conference on artificial intelligence. vol.~29 (2015)

\bibitem{rd2}
Niculescu-Mizil, A., Caruana, R.: Predicting good probabilities with supervised learning. In: Proceedings of the 22nd international conference on Machine learning. pp. 625--632 (2005)

\bibitem{SCEACE}
Nixon, J., Dusenberry, M.W., Zhang, L., Jerfel, G., Tran, D.: Measuring calibration in deep learning. In: CVPR workshops. vol.~2 (2019)

\bibitem{its}
Ouyang, C., Wang, S., Chen, C., Li, Z., Bai, W., Kainz, B., Rueckert, D.: Improved post-hoc probability calibration for out-of-domain mri segmentation. In: International Workshop on Uncertainty for Safe Utilization of Machine Learning in Medical Imaging. pp. 59--69. Springer (2022)

\bibitem{augmentation1}
Patel, K., Beluch, W., Zhang, D., Pfeiffer, M., Yang, B.: On-manifold adversarial data augmentation improves uncertainty calibration. In: 2020 25th International Conference on Pattern Recognition (ICPR). pp. 8029--8036. IEEE (2021)

\bibitem{ps}
Platt, J., et~al.: Probabilistic outputs for support vector machines and comparisons to regularized likelihood methods. Advances in large margin classifiers  \textbf{10}(3),  61--74 (1999)

\bibitem{kvasir}
Pogorelov, K., Randel, K.R., Griwodz, C., Eskeland, S.L., de~Lange, T., Johansen, D., Spampinato, C., Dang-Nguyen, D.T., Lux, M., Schmidt, P.T., et~al.: Kvasir: A multi-class image dataset for computer aided gastrointestinal disease detection. In: Proceedings of the 8th ACM on Multimedia Systems Conference. pp. 164--169 (2017)

\bibitem{etis}
Silva, J., Histace, A., Romain, O., Dray, X., Granado, B.: Toward embedded detection of polyps in wce images for early diagnosis of colorectal cancer. International journal of computer assisted radiology and surgery  \textbf{9},  283--293 (2014)

\bibitem{cvccol}
Tajbakhsh, N., Gurudu, S.R., Liang, J.: Automated polyp detection in colonoscopy videos using shape and context information. IEEE transactions on medical imaging  \textbf{35}(2),  630--644 (2015)

\bibitem{augmentation2}
Thulasidasan, S., Chennupati, G., Bilmes, J.A., Bhattacharya, T., Michalak, S.: On mixup training: Improved calibration and predictive uncertainty for deep neural networks. Advances in neural information processing systems  \textbf{32} (2019)

\bibitem{cvc300}
V{\'a}zquez, D., Bernal, J., S{\'a}nchez, F.J., Fern{\'a}ndez-Esparrach, G., L{\'o}pez, A.M., Romero, A., Drozdzal, M., Courville, A.: A benchmark for endoluminal scene segmentation of colonoscopy images. Journal of healthcare engineering  \textbf{2017} (2017)

\bibitem{ets}
Zhang, J., Kailkhura, B., Han, T.Y.J.: Mix-n-match: Ensemble and compositional methods for uncertainty calibration in deep learning. In: International conference on machine learning. pp. 11117--11128. PMLR (2020)

\bibitem{areview}
Zou, K., Chen, Z., Yuan, X., Shen, X., Wang, M., Fu, H.: A review of uncertainty estimation and its application in medical imaging. Meta-Radiology p. 100003 (2023)

\end{thebibliography}

\end{document}